\documentclass[10pt,twocolumn,letterpaper]{article}

\usepackage{iccv}      
\usepackage{multirow}
\usepackage{multicol}
\usepackage{colortbl}
\usepackage{xcolor}  
\definecolor{darkgreen}{rgb}{0.0, 0.5, 0.0}
\usepackage{graphicx}
\usepackage{subcaption}
\usepackage{caption} 
\usepackage{makecell}
\usepackage{siunitx}

\usepackage{epstopdf}

\definecolor{iccvblue}{rgb}{0.21,0.49,0.74}
\usepackage[pagebackref,breaklinks,colorlinks,allcolors=iccvblue]{hyperref}

\def\paperID{156} 
\def\confName{ICCV}
\def\confYear{2025}

\title{CalliReader\includegraphics[width=0.65cm]{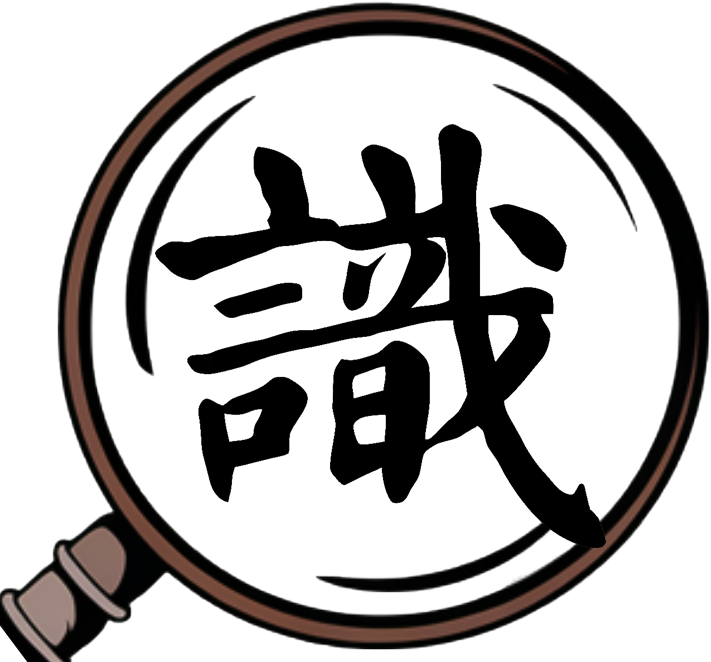}: Contextualizing Chinese Calligraphy via \\an Embedding-Aligned Vision-Language Model}

\vspace{-25mm}
\author{Yuxuan Luo$^{1}$\thanks{denotes equal contribution},~~Jiaqi Tang$^{1}$\footnotemark[1],~~Chenyi Huang$^{1}$,~~Feiyang Hao$^{2}$,~~Zhouhui Lian$^{1}$\footnotemark[2]\thanks{denotes corresponding author}\\
    $^1$Wangxuan Institute of Computer Technology, Peking University\\
    $^2$Xi’an Jiaotong University 
     \\
    \url{https://github.com/LoYuXr/CalliReader}
}

\begin{document}

\twocolumn[{%
\renewcommand\twocolumn[1][]{#1}%
\maketitle
\vspace{-2.9em}
\includegraphics[width=0.99\linewidth]{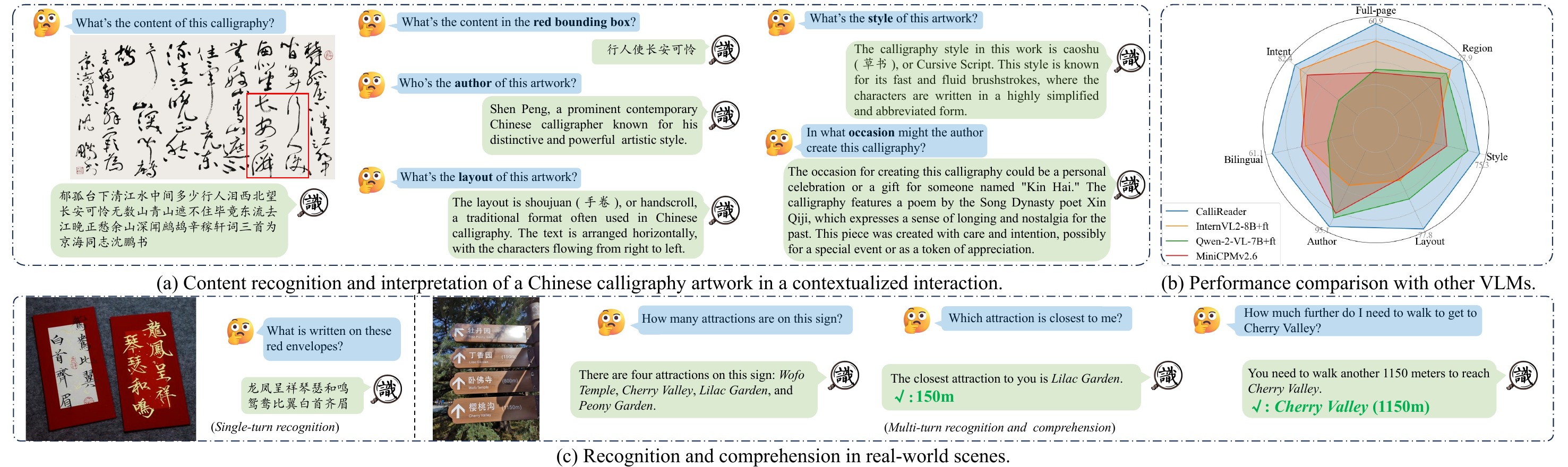}
\centering
\vspace{-1em}
\captionof{figure}{We propose \textit{CalliReader}, a versatile vision-language model (VLM) for Chinese Calligraphy Contextualization (CC$^2$). It excels in full-page/region-wise recognition, multilingual interpretation, and intent identification of calligraphy artworks (a), outperforming fine-tuned state-of-the-art VLMs (b), while also demonstrating applicability in general scene-text recognition and comprehension (c).  } 
\label{fig:teaser}}
\vspace{0.2em}]
\begin{abstract}
Chinese calligraphy, a UNESCO Heritage, remains computationally challenging due to visual ambiguity and cultural complexity. Existing AI systems fail to contextualize their intricate scripts, because of limited annotated data and poor visual-semantic alignment. We propose CalliReader, a vision-language model (VLM) that solves the Chinese Calligraphy Contextualization (CC$^2$) problem through three innovations: (1) character-wise slicing for precise character extraction and sorting, (2) CalliAlign for visual-text token compression and alignment, (3) embedding instruction tuning (e-IT) for improving alignment and addressing data scarcity. We also build CalliBench, the first benchmark for full-page calligraphic contextualization, addressing three critical issues in previous OCR and VQA approaches: fragmented context, shallow reasoning, and hallucination. Extensive experiments including user studies have been conducted to verify our CalliReader's \textbf{superiority to other state-of-the-art methods and even human professionals in page-level calligraphy recognition and interpretation}, achieving higher accuracy while reducing hallucination. Comparisons with reasoning models highlight the importance of accurate recognition as a prerequisite for reliable comprehension. Quantitative analyses validate CalliReader's efficiency; evaluations on document and real-world benchmarks confirm its robust generalization ability.

\end{abstract}

\vspace{-4mm}
\section{Introduction}
\label{sec:intro}
\vspace{-1mm}
Chinese calligraphy is recognized by UNESCO as an Intangible Cultural Heritage. It possesses intricate scripts, arbitrary layouts, and historical glyph variations, making it hard to read and understand, even for native Chinese speakers. Yet, calligraphy encapsulates immense cultural and historical significance. Therefore, unlocking its rich context through effective Chinese Calligraphy Contextualization (CC²) methods is paramount. Current systems struggle to achieve this, requiring \textbf{precise recognition, in-depth comprehension, and multilingual interpretation.}

 
Existing optical character recognition (OCR) tools~\cite{PaddleOCR, AliOCR} falter when faced with calligraphy's stylistic nuances. Vision-language models (VLMs) offer a promising alternative, demonstrating proficiency in image understanding and interpretation~\cite{internlmxcomposer, internlmxcomposer2_5, yao2024minicpm}. However, VLMs often lack training on datasets tailored to calligraphy. Additionally, subtle character variations in Chinese calligraphy, such as near-identical characters, necessitate high-resolution inputs. 
             
In this paper, we aim to leverage the rich visual semantics brought by pre-trained vision-transformer (ViT)-based encoders~\cite{dosovitskiy2021an}. By introducing plug-and-play modules, we reorganize visual input representations and perform semantic compression and alignment. In this manner, we reduce the training burden while maintaining VLMs' capabilities.

To this end, we propose \textit{CalliReader}, a framework designed to enhance CC$^2$ capabilities. It plugs two modules to a base VLM (e.g., InternVL2-8B~\cite{chen2024far}): character-wise slicing and \textit{CalliAlign}. The slicing process extracts individual characters from pages, simplifying the recognition task while preserving character semantics. \textit{CalliAlign} compresses character visual tokens and aligns them with normalized text embeddings, enabling performance scaling on large single-character datasets. This design preserves ViT's original representation. Pilot experiments validate the effectiveness of character slicing and the potential of \textit{CalliAlign} for improved visual-to-knowledge inference on CC$^2$ tasks.

We further design embedding instruction-tuning (e-IT) to combat data scarcity. It unifies authentic and synthetic data within the same embedding representation. This allows augmentation and enables efficient adaptation of the LLM via LoRA~\cite{hu2021lora}. Instructed with page-level recognition, e-IT-equipped \textit{CalliReader} surpasses LoRA fine-tuned VLMs on comprehension and interpretation. This highlights the efficacy of e-IT in mining contextual information.

We also build a high-quality page-level calligraphy dataset, comprising 7,357 training and 3,192 test samples. Using the test set, we develop CalliBench, a comprehensive benchmark centered on CC$^2$. It features full-page recognition, regional hallucination detection, and contextual VQA for knowledge grounding, bilingual interpretation, and intent analysis. \textbf{CalliBench provides a text-centric recognition and cultural-axis multimodal reasoning benchmark for the community}, addressing the fragmented context and shallow reasoning in OCR and VQA benchmarks.

Comparisons with representative fine-tuned VLMs and prevalent OCR models on CalliBench demonstrate \textit{CalliReader}'s superiority. It also outperforms state-of-the-art reasoning models, enlightening that versatile reasoning must be grounded in precise recognition. \textbf{Surprisingly, our user study demonstrates that \textit{CalliReader} remarkably surpasses not only the average level of native speakers but also that of calligraphy experts.} Further applications on document, handwriting, and in-the-wild text-centric visual benchmarks validate its generalization ability, paving the way for broad downstream applications.
\vspace{-3mm}
\section{Related Work}
\label{sec:related}
\vspace{-3mm}
\textbf{Vision-language models (VLMs).} The proposal of scaling laws~\cite{kaplan2020scaling, hoffmann2022training} and novel training techniques~\cite{radford2018GPT, brown2020GPT3,christiano2017deep,ouyang2022instructGPT} has driven the rapid proliferation of decoder-only language models, exhibiting superior instruction-following and few-shot learning capabilities~\cite{openai2022chatgpt, chowdhery2023palm, achiam2023gpt4, zhang2022opt,touvron2023llama,touvron2023llama-2, jiang2023mistral, du2021glm}. Concurrently, smaller foundation models trained on large-scale, high-quality data have also gained prominence~\cite{cai2024internlm2, yang2023baichuan, hu2024minicpm}. The advent of ViT~\cite{dosovitskiy2021an} and CLIP~\cite{radford2021CLIP} has empowered VLMs to process multimodal inputs within a unified embedding space~\cite{liu2024Llava, liu2024llava1_5, li2023blip-2, Qwen-VL, Qwen2VL, xu2024llava-uhd, liu2024textmonkey}. Current research emphasizes high-resolution tasks, where scaled vision backbones improve both visual resolution and task generalization~\cite{jia2021scaling, fang2023eva, chen2023internvl(vit), yang2024vision, li2022blip}. Additionally, slicing strategies extend the model to handle up to 4K resolutions~\cite{liu2024llava1_5,xu2024llava-uhd,chen2024far}. However, despite notable success on OCR benchmarks~\cite{mathew2021docvqa,masry2022chartqa,mathew2021infographicvqa,singh2019towards,ocrbench2023}, VLMs still exhibit severe hallucinations in visual contextualization, especially in low-resource scenes like Chinese calligraphy. For CC$^2$, current VLMs heavily rely on memorization rather than comprehension, thus suffering from hallucinations.


\noindent\textbf{Vision to Knowledge.} Conventional OCR methods focus on extraction and detection~\cite{zhong2019publaynet,tong2020ma,lyu2018multi,liao2017textboxes,liao2022real, peng2022pagenet,wang2020contournet} but lack depth in knowledge extraction. OCR-based LLMs for document VQA~\cite{lu2024bounding,liu2024textmonkey,wei2024general} struggle with stylistic layout variations, especially in Chinese calligraphy. Modern reasoning models, despite CoT capabilities~\cite{ji2024align,qvq-72b-preview,OpenAI2024}, often hallucinate due to ambiguous recognition. 

\noindent\textbf{Existing Text-Centric Vision Benchmarks}~\cite{zhang2020scut,mthv2,singh2019towards,mathew2021docvqa} primarily focus on standard text spotting or shallow Q\&A. Most visual-reasoning~\cite{chen2024spatialvlm,cheng2024spatialrgpt} emphasize spatial and causal relations, leaving historical and cultural analysis unexplored. \textit{CalliBench} emphasizes full-context recognition and interpretation. With accurate recognition of stylistic content and multi-faceted analysis, \textit{CalliBench} sets a new standard for evaluating VLM's ability to perform flexible, knowledge-intensive reasoning in culturally rich contexts.

\noindent\textbf{Visual compression, enhancement, and alignment.} Multimodal integration causes additional computation and misalignment, emphasizing effective compression, enhancement, and cross-modal alignment. MLP variants~\cite{liu2024Llava, liu2024llava1_5, li2024llava-next, lin2023sphinx, internlmxcomposer, internlmxcomposer2, internlmxcomposer2_5} are commonly used. Learnable-query-based transformers~\cite{alayrac2022flamingo, li2023blip-2, ye2023mplug, hu2024minicpm, Qwen-VL, Qwen2VL, li2024monkey, huang2024minimonkey, liu2024textmonkey} draw inspiration from object detection~\cite{carion2020end}, and can satisfy above requirements. Non-parametric clustering~\cite{ jin2024chat, wu2024towards}, bi-partie graph~\cite{bolya2023token}, token replacement~\cite{huang2024minimonkey, shang2024llavapru} and adaptive pooling~\cite{yao2024deco} preserve multi-granularity visual semantics. Drawing inspiration from these approaches~\cite{zhao2024mg-llava, zhang2024omg-llava}, we develop a parametric character-wise slicing and adopt the perceiver resampler~\cite{alayrac2022flamingo} architecture.

\vspace{-2mm}
\section{Pilot Experiments}
\vspace{-2mm}
\label{sec:pioneer}
The two pilot experiments examine: (1) the optimal slicing strategy for identifying visual characters, (2) VLM's tolerance to noisy inputs. The first leads to our character-wise, semantic-preserved slicing, while the second ensures LLM is compatible with pseudo-text tokens from \textit{CalliAlign}.
\vspace{-1.5mm}
\subsection{Optimal Slicing Strategy}
\vspace{-1.5mm}
\label{sec:slicing}
VLMs use slicing to decompose high-resolution images, but brute-force cropping may separate character radicals. Character-wise slicing enhances the model's recognition and reasoning in the CC$^2$ task.

\begin{figure}[t]
    \centering
    \includegraphics[width=\linewidth]{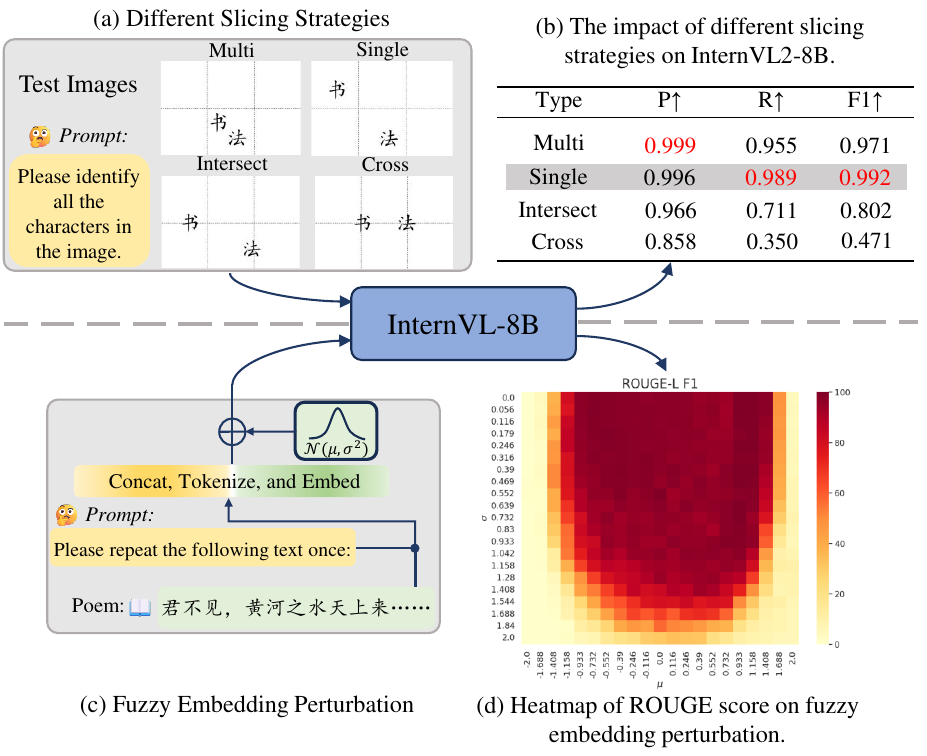}
    \vspace{-7mm}
    \caption{Pilot experiments. (a): Finding the optimal slicing 
    strategy; (b): Single is the best slicing policy; (c): Fuzzy inputs test the LLM's deviation; (d): 
    The heatmap depicts the degradation limit.}

    \label{fig:pilot-experiments}
\vspace{-4mm}
\end{figure}

We evaluate various slicing and their impact on recognition accuracy. Fig.~\ref{fig:pilot-experiments} shows four layouts: (1) \textbf{Multi}: multiple characters per slice, (2) \textbf{Single}: one character per slice, (3) \textbf{Intersect}: characters spanning two slices, and (4) \textbf{Cross}: characters divided into four quadrants. We generate 200 images for each layout, with 6 random characters rendered in the Sung typeface. Using InternVL2 with its 448$\times$448 ViT, we embed the image and ask to identify all written characters. We calculate average precision, recall, and macro F1 scores. Fig.~\ref{fig:pilot-experiments} (b) shows that the Single layout achieves the highest recall and F1 scores.

\vspace{-2mm}
\subsection{Fuzzy Embeddings for LLMs}
\label{sec:fuzzy}
\vspace{-1mm}

Previous VLMs integrate multimodal inputs within the embedding space, yet few provide quantitative analysis of their tolerance to ambiguous inputs. To validate \textit{CalliAlign}'s ability to generate pseudo-text embeddings from character images, we introduce Gaussian noise to perturb text embeddings and analyze the base LLM’s deviation.

In Fig.~\ref{fig:pilot-experiments} (c), the model is tasked to repeat 100 classical Chinese sentences from the Erya dataset~\cite{guo2023towards}, with Gaussian noise $\mathcal{N}(\mu, \sigma^2), \mu \in [-2, 2], \sigma \in [0, 2]$ applied to layer-normalized text embeddings. Output fidelity is measured using the ROUGE-L score. Fig.~\ref{fig:pilot-experiments} (d) shows stable LLM outputs when $\mu \in [-1, 1]$ and $\sigma \in [0, 1.2]$, with significant degradation outside these ranges. These findings characterize LLM's tolerance and guide \textit{CalliAlign} training.

\vspace{-7mm}
\section{Method}
\begin{figure}[t]
    \centering
    \includegraphics[width=\linewidth]{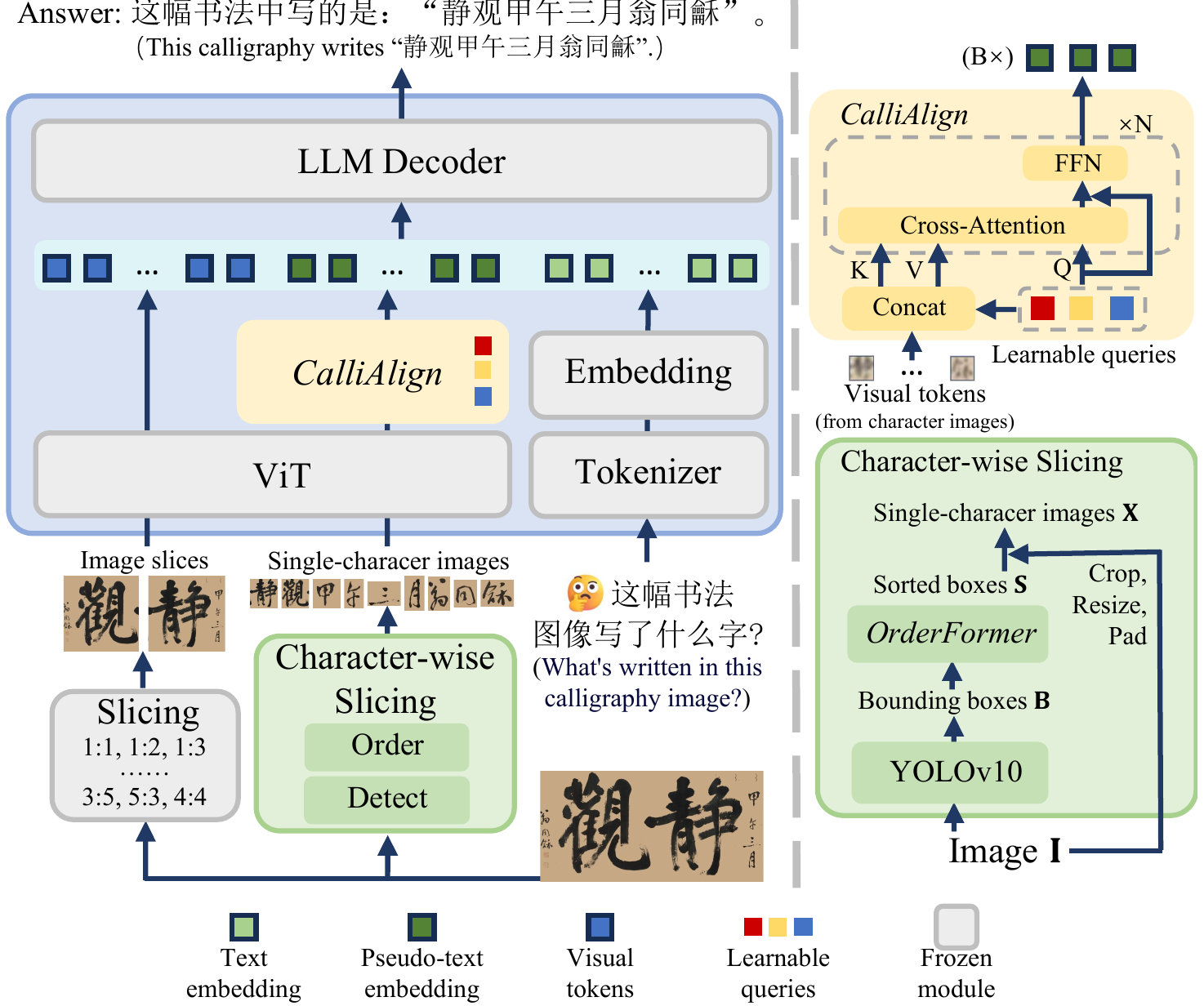}
    \vspace{-6mm}
    \caption{(Left) An overview of \textit{CalliReader} featuring pre-trained Slicing and \textit{CalliAlign} modules. (Right) Detailed architecture of the pluggable modules. \textit{CalliAlign} uses a perceiver-resampler, and the slicing process adopts YOLOv10 and \textit{OrderFormer}.}
    \label{fig:overall_model}
    \vspace{-4mm}
\end{figure}

\vspace{-2mm}
Fig.~\ref{fig:overall_model} shows our approach, with three core innovations:
\begin{enumerate}
    \item \textbf{Character-wise Slicing} uses detection and sorting modules, reducing page-level analysis to ordered character-level recognition.

    \item \textbf{\textit{CalliAlign}} processes ViT-extracted character features and maps them to pseudo-text embeddings. Trained and scaled on single-character datasets, it achieves accurate alignment and reduces sequence length.
    \item \textbf{Embedding Instruction Tuning} (e-IT) unifies authentic and synthesized data in text embeddings, applying LoRA fine-tuning. This alleviates data insufficiency, tailors the LLM to \textit{CalliAlign}, and reduces hallucinations.
\end{enumerate}
These pluggable modules improve CC$^2$ accuracy with only 0.88B additional parameters ($10\%$ of an 8B VLM).

\begin{figure*}[t]
    \vspace{-0mm}
    \centering
    \includegraphics[width=\linewidth]{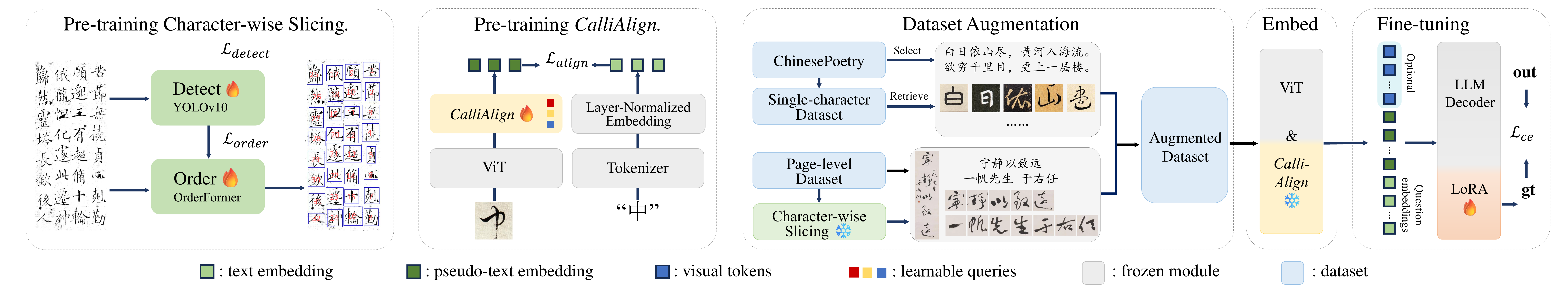}
    \vspace{-7mm}
    \caption{Training \textit{CalliReader} modules. Left: pre-training character-wise slicing modules on the page-level dataset. Mid: pre-training \textit{CalliAlign} solely on our single-character dataset, fitting pseudo-text tokens to padded, layer-normalized text embeddings. Right: embedding instruction tuning (e-IT), where we preprocess and augment training embeddings and implement LoRA on InternLM2.5‑7b‑chat.}
    \label{fig:wholemodel}
    \vspace{-5mm}
\end{figure*}

\vspace{-2mm}
\subsection{Character-wise Slicing}
\vspace{-1mm}
Motivated by Section~\ref{sec:slicing}, we develop character-wise slicing to preserve layout information. It identifies character regions, extracts ordered image patches and preserves semantic information for the VLM.

Unlike adaptive slicing~\cite{chen2024far}, which searches for optimal aspect ratios, we fine-tune a YOLOv10~\cite{wang2024yolov10} for bounding-box detection and develop an \textit{OrderFormer} encoder for ordering. Fig.~\ref{fig:wholemodel} (Left) shows the training details. Both modules are trained on page-level data using YOLOv10's default detection loss $\mathcal{L}_{detect}$, and MSE loss $\mathcal{L}_{order}$. These modules contain 0.31B parameters in total, with implementation details provided in the supplementary material.

During inference (Fig.~\ref{fig:overall_model}), a high-resolution image $\mathbf{I} \in \mathbb{R}^{H \times W \times 3}$ is processed by Detect and Order modules. YOLOv10 $f_{\text{y}}$ first identifies character bounding-boxes $\mathbf{B}$, and \textit{OrderFormer} $f_{\text{o}}$ recovers the reading sequence $\mathbf{S}$:
\vspace{-2.2mm}
\begin{equation}
  \begin{aligned}
    \mathbf{B} &= f_{\text{y}}(\mathbf{I}) = \{B_{i}\}, \quad
    \mathbf{S} = f_{\text{o}}(\mathbf{B}) = \{S_{i}\}, \\
    B_i&=(x_{i_1},y_{i_1},x_{i_2},y_{i_2}),\quad i = 1,2\dots n.
  \end{aligned}
  \vspace{-1.2mm}
\end{equation} The image character sequence $\mathbf{X} \in \mathbb{R}^{n \times h \times w \times 3}$ is obtained by cropping, padding, and resizing:
\vspace{-1.5mm}
\begin{equation}
    \mathbf{X} = \text{Resize}(\text{Pad}(\text{Crop}(\mathbf{S}, \mathbf{I})), h \times w).
    \vspace{-1.5mm}
\end{equation}

The above process preserves order and semantics, enhancing subsequent Chinese calligraphy-centric reasoning.
\vspace{-7.2mm}
\subsection{\textbf{\textit{CalliAlign}}}
\vspace{-1.5mm}

The image character sequence $\mathbf{X}$ encodes semantic and order information. However, excessive sequence length may exceed the LLM capacity. Instead of directly using ViT token sequences, \textit{CalliAlign} maps visual tokens to text token `labels', enhancing reasoning and efficiency.

Based on the perceiver resampler~\cite{alayrac2022flamingo,ye2023mplug}, the 0.57B \textit{CalliAlign} projects 256 visual tokens into 3 pseudo-text embeddings (for each character), preserving authentic, normalized Chinese character embeddings. This achieves $98.8\%$ compression, extends sequence limits, and reduces LLM training and inference overhead:
\vspace{-2mm}
\begin{equation}
    \vspace{-1mm}
    \mathbf{T}^{'}_c = CalliAlign(f_V(\mathbf{X})) \in \mathbb{R}^{n \times q \times d},
    \vspace{-1mm}
\end{equation}
where ViT $f_V$ extracts features from $\textbf{X}$, and \textit{CalliAlign} generates pseudo-text embeddings $\textbf{T}^{'}_c$ (Fig.~\ref{fig:overall_model}).

Furthermore, to reduce embedding variance~\cite{zeng2023glmb,openai_embeddings_guide} and stabilize training, we layer-normalize VLM's textual embeddings $\textbf{T}$ into $\textbf{T}'$ as ground truth:
\vspace{-2.5mm}
\begin{equation}
    \mathbf{T}^{'} = \frac{\mathbf{T}-\mu}{\sigma},
    \vspace{-2.5mm}
\end{equation}
where $\mu$ and $\sigma$ represent the mean and standard deviation. 

\textit{CalliAlign} simplifies the training process while preserving the original vision backbone. During training (illustrates in  Fig.~\ref{fig:wholemodel}), we apply L2 loss $\mathcal{L}_{align}$ to $(\textbf{X},\textbf{T}')$ pairs. This enables efficient scaling using a sufficient single-character image dataset. In inference, $\textbf{T}^{'}_c$ is denormalized before LLM integration.  
\vspace{-1.5mm}
\subsection{Embedding Instruction-tuning}
\label{sec:embedinstructtuning}
\vspace{-1.5mm}
Conventional LLM fine-tuning methods modify fusion layers~\cite{Qwen-VL,lin2023sphinx}, the LLM~\cite{ye2023mplug,internlmxcomposer2}, or both~\cite{li2024monkey,xu2024llava-uhd,zhao2024harmonizing}, requiring extensive data and thus being unsuitable for limited page-level data scenarios. To address this, we propose Embedding Instruction-Tuning (e-IT), leveraging abundant single-character images to augment training data.

e-IT unifies synthetic and authentic instruction-tuning data within a shared embedding space. The format concatenates embeddings: tokenized user queries + \textit{CalliAlign} pseudo-text embeddings. When available, global visual features from page-level images are appended. This unified embedding format enables augmentation with single-character images and textual corpora, reducing reliance on extensive page-level annotations.

For augmentation, we create single-character sequence queries from ChinesePoetry~\cite{chinese-poetry} contents, slice page-level images, and encode them using \textit{CalliAlign} (Fig.~\ref{fig:wholemodel}, Right). Only LLM is fine-tuned to follow user instructions, recognizing, interpreting, and inferring knowledge from the original content. e-IT adapts the LLM to \textit{CalliAlign}, leveraging internal knowledge for error correction and discovery. e-IT enables scalable LoRA fine-tuning, reducing memory requirements while improving performance.

\vspace{-2mm}
\section{Dataset and Benchmark}
\vspace{-2mm}
This section details our collected single-character and page-level datasets. From the latter, we develop a comprehensive CC$^2$ benchmark, named CalliBench.
\vspace{-1.5mm}
\subsection{Single-Character dataset}
\vspace{-1.5mm}
\label{sec:datasc}
We collect 742,975 stylized calligraphy character images from Shufazidian.com~\cite{shufazidian}, covering 6,763 GB2312 standard characters. We split 520,083 for training, 111,446 for validation, and 111,446 for test samples to train \textit{CalliAlign}.
\vspace{-7mm}
\subsection{Page-Level dataset} 
\vspace{-2mm}
\label{sec:datapl}
We curate 10,549 densely annotated page-level calligraphy images from ArtronNet~\cite{artronnet} and CAOD~\cite{CAOD}, featuring diverse styles and layouts. 7,357 samples are used for training, and 3,192 constructs CalliBench. Character bounding boxes and contents are manually labeled using the LabelMe format. The dataset trains character-wise slicing and fine-tunes the LLM. For e-IT, we randomly select 6,000 poems from the ChinesePoetry corpus~\cite{chinese-poetry} for augmentation. Details of our datasets can be found in our supplementary.

\vspace{-1mm}
\subsection{CalliBench: Multi-Level CC$^2$ Evaluation}
\vspace{-1mm}
\label{sec:callibench}

\begin{figure*}[t]
    \centering
    \includegraphics[width=\linewidth]{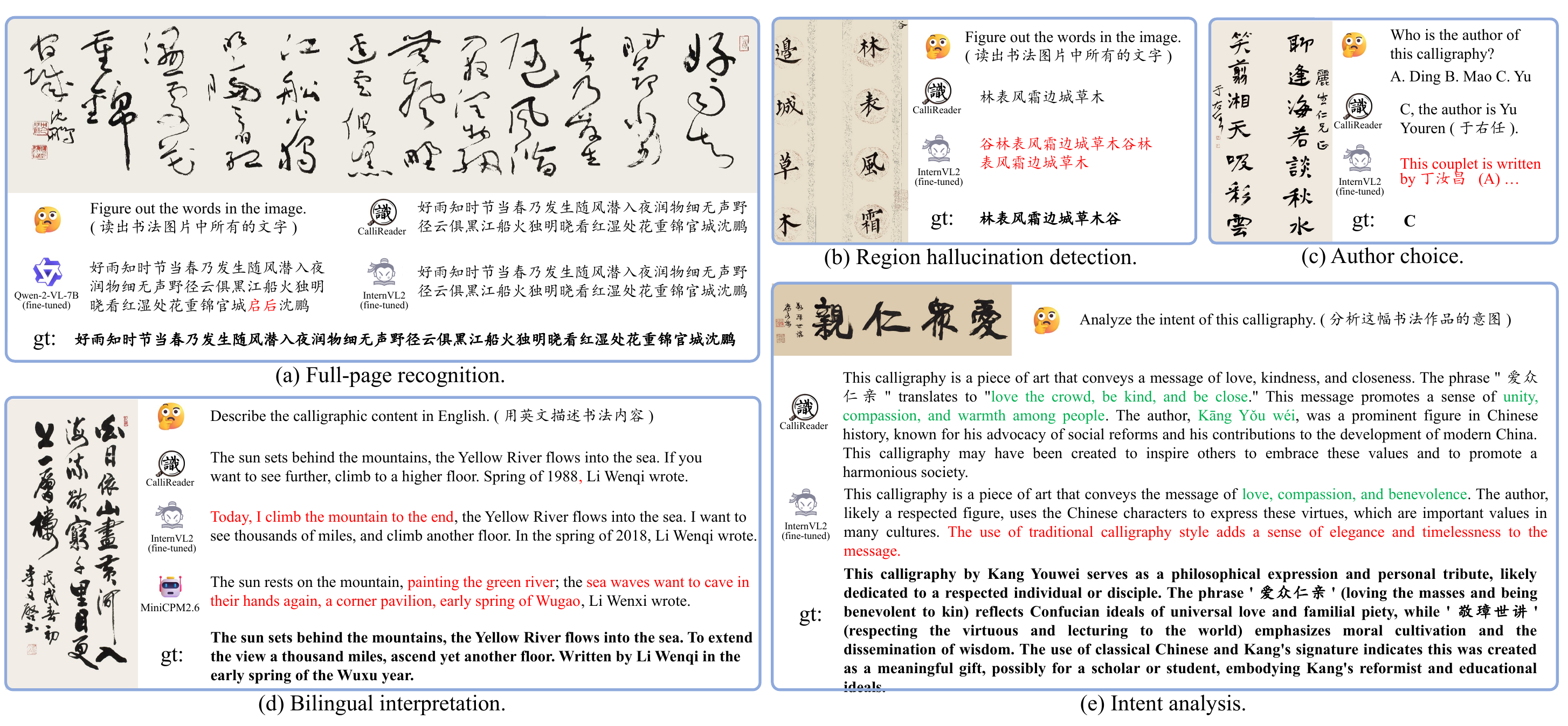}
    \vspace{-7mm}
    \caption{Visualization of CalliBench. We present the performance of CalliReader and other models across three tasks: Full-page Recognition (Top Left), Regional Hallucination Detection (Top Center), and Contextual VQA (Top Right, Bottom). Errors are colored in red.}
    \label{fig:callibench_vis}
    \vspace{-5mm}
\end{figure*}

CalliBench evaluates calligraphic contextualization at multiple granularities. It evaluates full-page recognition, regional hallucination detection, and contextual VQA through OCR, multiple-choice, and open-ended questions. Table~\ref{fig:callibench} and Fig.~\ref{fig:callibench_vis} provide visual format and examples.

\begin{table}[t!]
\centering
    \includegraphics[width=\linewidth]{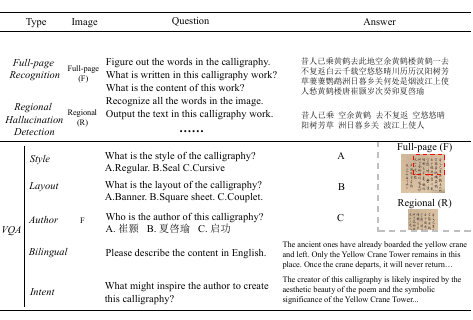}
    \vspace{-8mm}
    \caption{ CalliBench format, evaluating recognition, hallucination resistance, and contextual VQA for knowledge grounding.}
    \label{fig:callibench}
    \vspace{-4mm}
\end{table}

\begin{table*}[t!]
    \centering
    \begin{small}
    \vspace{-3mm}
    \begin{tabular}{l|l|cccc|cccc|cccc}
        \hline
        \multirow{2}{*}{Model} & \multirow{2}{*}{Type} & \multicolumn{4}{c|}{Easy} & \multicolumn{4}{c|}{Medium} & \multicolumn{4}{c}{\textbf{Hard}}\\ \cline{3-14} 
         &  & P↑ & R↑ & F1↑ & NED↓ & P↑ & R↑ & F1↑ & NED↓ & P↑ & R↑ & F1↑ & NED↓ \\ \hline
        EasyOCR~\cite{EasyOCR} & OCR & 0.341 & 0.212 & 0.253 & 0.941 & 0.185 & 0.088 & 0.112 & 0.971  & 0.105 & 0.031 & 0.038 & 0.991\\
        Ali-OCR~\cite{AliOCR} & OCR & 0.449 & 0.284 & 0.330 & 0.888 & 0.305 & 0.143 & 0.183 & 0.928 & 0.264 &0.102 & 0.128 & 0.947\\
        PP-OCRv4~\cite{PaddleOCR} & OCR & 0.632 & 0.513 & 0.554 & -- & 0.487 & 0.319& 0.368 & -- & 0.368 & 0.177 & 0.220 & --\\
        YOLOv10~\cite{wang2024yolov10} & OCR & 0.846 & 0.774 & 0.801 & -- & 0.765 & 0.612 & 0.668 & -- & 0.641 & 0.324 & 0.388 & --\\ \hline
        LLAVA-NEXT~\cite{li2024llava-next} & VLM & 0.022 & 0.052 & 0.026 & 0.990 & 0.034 & 0.057 & 0.036 & 0.988 & 0.072 & 0.060 & 0.051 & 0.984\\
        MiniCPM2~\cite{yao2024minicpm} & VLM & 0.211 & 0.165 & 0.156 & 0.952 & 0.146 & 0.091 & 0.091 & 0.974 & 0.142 & 0.050 & 0.054 & 0.984\\ 
        MiniCPM2.6~\cite{yao2024minicpm} & VLM & 0.256 & 0.470 & 0.288 & 0.867 & 0.201 & 0.318 & 0.217 & 0.905 & 0.150 & 0.169 & 0.132 & 0.953\\ 
        Qwen2-VL-7B~\cite{Qwen-VL} & VLM & 0.573 & 0.506 & 0.502 & 0.708 & 0.456 & 0.348 & 0.367 & 0.786 & 0.304 & 0.170 & 0.191 & 0.904\\
        Qwen-VL-Max~\cite{Qwen-VL-max} & VLM & 0.582 & 0.480 & 0.511 & 0.641 & 0.517 & 0.394 & 0.429 & 0.681 & 0.326 & 0.250 & 0.246 &0.857\\
        Qwen2.5-VL-7B~\cite{qwen2025qwen25technicalreport} & VLM & 0.440 & 0.736 & 0.534 & 0.710 & 0.401 & 0.562 & 0.455 & 0.737 & 0.280 & 0.325 & 0.278 & 0.863\\
        GPT-4o~\cite{GPT-4o} & VLM & 0.457 & 0.403 & 0.400 & 0.726 & 0.301 & 0.244 & 0.252 & 0.840 & 0.174 & 0.102 & 0.101 & 0.954\\
        MiniMonkey~\cite{huang2024minimonkey} & VLM & 0.642 & 0.621 & 0.605 & 0.641 & 0.550 & 0.521 & 0.510 & 0.665 & 0.347 & 0.318 & 0.303 & 0.807\\ 
        GOT-OCR2.0~\cite{wei2024general} & VLM & 0.687 & 0.550 & 0.593 & 0.651 & 0.508 & 0.334 & 0.373 & 0.765 & 0.375 & 0.135 & 0.152 & 0.936\\ 
        InternVL2-8B~\cite{chen2024far} & VLM & 0.729 & 0.598 & 0.617 & 0.631 & 0.699 & 0.587 & 0.625 & 0.603 & 0.415 & 0.314 & \cellcolor{red!20}0.324 & 0.803\\ \hline 
        \rowcolor{lightgray}CalliReader w/o e-IT & VLM & 0.881 & 0.721 & 0.779 & 0.344 & 0.698 & 0.675 & 0.677 & 0.445 & 0.543 & 0.341 & 0.390 & 0.745\\ \hline
        MiniCPM2.6+ft & LoRA & 0.756 & 0.759 & 0.757 & 0.309 & 0.554 & 0.555 & 0.552 & 0.537 & 0.351 & 0.322 & 0.329 & 0.790\\
        Qwen2-VL-7B+ft & LoRA & 0.789 & 0.791 & 0.789 & 0.280 & 0.663 & 0.528 & 0.587 & 0.476 & 0.367 & 0.373 & 0.348 & 0.757\\
        Qwen2.5-VL-7B+ft & LoRA & 0.886 & 0.879 & 0.881 & 0.150 & 0.729 & 0.723 & 0.724 & 0.510 & 0.490 & 0.494 & 0.474 & 0.625\\
        InternVL2-8B+ft & LoRA & 0.900 & 0.895 & 0.896 & 0.130 & 0.798 & 0.787 & 0.789 & 0.255 & 0.530 & 0.534 & \cellcolor{red!20}0.511 &0.579\\
        \hline
         \rowcolor{lightgray}CalliReader & LoRA & \textbf{0.912} & \textbf{0.903} & \textbf{0.907} & \textbf{0.121} & \textbf{0.822} & \textbf{0.807} & \textbf{0.813} & \textbf{0.232} & \textbf{0.637} & \textbf{0.593} & \textbf{0.609} & \textbf{0.516}\\ 
        
        \hline
    \end{tabular}
    \end{small}
    \vspace{-3mm}
    \caption{\textbf{Full-page Recognition performance.} Metrics include average precision (P), recall (R), Macro-F1 (F1), and normalized edit distance (NED) across all tiers. Results show \textit{CalliReader}'s superior performance compared to other off-the-shelf VLMs and OCR models. On the Hard tier, F1 scores of InternVL2-8B and the fine-tuned variant are marked in red, highlighting our model's improvement.}
    \label{tab:full_page_ocr}
    \vspace{-3.5mm}
\end{table*}

\noindent\textbf{Full-page Recognition} evaluates the model's precise visual OCR capabilities. The test set includes three tiers: easy (727 samples; regular script; fewer characters), medium (2,212 samples), and hard (253 samples; cursive scripts; diverse layouts). We evaluate models using eight content-centered questions, measuring average precision (P), recall (R), and normalized edit distance (NED).

\noindent\textbf{Region Hallucination Detection} evaluates model resistance to hallucinations. Incomplete content may cause VLMs to generate redundant associations or outputs, guessing or reciting rather than truly recognizing. We generate 727 test samples by randomly cropping easy-level images to include partial content regions.

\noindent\textbf{Contextual VQA} evaluates model comprehension and interpretation of calligraphy background knowledge. A multi-turn vision-Q\&A format begins with full-page recognition queries, followed by knowledge-based questions. Questions probe understanding of calligraphic style, layout, authorship, bilingual interpretation, and intent analysis:
\begin{itemize}

    \item \textbf{Multiple-Choice Questions} evaluate style, layout, and authorship. Models are required to select one correct answer from three options, and their responses are assessed through accuracy (Acc). For style and layout, we select 750 and 806 images, respectively, from all tiers. Additionally, we manually annotate a set of 1,194 images attributed to 139 authors for the authorship questions.


    \item \textbf{Bilingual Interpretation} evaluates content rephrasing. In the second stage (as visualized in Table~\ref{fig:callibench}), we assess English interpretation capabilities. We verify the outputs against the ground truth using Sentence Transformer~\cite{reimers-2019-sentence-bert} and calculate the cosine similarity (STSim). The test set consists of 500 calligraphy-translation pairs from the medium tier, which have been manually annotated.

    \item \textbf{Intent Analysis} aims to discern the motivations behind calligraphic artwork (e.g., commemoration, archival, or decorative purposes). This helps reveal calligraphic passages' historical and cultural significance, providing deeper insights into their context. We randomly select 500 samples and evaluate the models' responses using top-performing LLMs such as DeepSeek-V3~\cite{deepseekai2024deepseekv3technicalreport} and Qwen2.5-Max~\cite{qwen2025qwen25technicalreport}, guided by a well-designed prompt. The Calligraphic Intent Score (CIS) ranges from 0 to 10 and is then multiplied by 10. Details of the evaluation framework can be found in the supplementary material.
    
\end{itemize}

\vspace{-3mm}
\section{Experiments}
\vspace{-1mm}
\subsection{Implementation Details}
\vspace{-1mm}
\textit{CalliReader} builds on InternVL2-8B~\cite{chen2024far}, combining InternViT-300M-448px and InternLM2.5‑7b‑chat, for state-of-the-art recognition and comprehension.

We use YOLOv10~\cite{wang2024yolov10} for bounding box detection, trained for 1,000 epochs on 8$\times$RTX4080 GPUs with page-level data, a learning rate (lr) of 1e-2, a batch size of 80, and SGD optimization. \textit{OrderFormer} is a 4-layer encoder-only Transformer with a maximum sequence length of 50, 8 attention heads per layer, 256 feature dimensions, and 4/1 input/output dimensions. We augment 57,627 column-wise ordering samples, training with a batch size of 4, AdamW optimizer (1e-2 learning rate), and CosineAnnealing with a warmup scheduler for 1000 epochs on an RTX4080 GPU.

\textit{CalliAlign} consists of 4 layers with 3 learnable queries, 64 attention heads, and a 4096-dimensional feature space. The training uses 4×A6000 GPUs for 50,000 steps on the single-character dataset, with a 1e-4 learning rate, batch size of 256, AdamW optimizer, and CosineAnnealing scheduler. Slicing and aligning modules are individually trained before integration to enhance VLM CC$^2$ capabilities.

For fine-tuning, we create the instruction dataset on full-page recognition format since manually annotating intention and interpretation sets is highly costly. We process 7,357 training pages and augment an extra 6,000 for e-IT. We apply e-IT to InternLM2.5-7b-chat using XTuner~\cite{2023xtuner} on 2$\times$A6000 GPUs with batch size 2 and deepspeed ZeRO-1 for 1 epoch, following Section~\ref{sec:embedinstructtuning}'s embedding instruction-tuning.
\vspace{-2mm}
\subsection{Full-Page Recognition}
\vspace{-2mm}
We evaluate \textit{CalliReader}'s full-page recognition performance with precision (P), recall (R), macro-F1 (F1), and normalized edit distance (NED). The evaluation includes top-performing open-source VLM models (LlaVA-NEXT~\cite{li2024llava-next}, MiniCPM-Vs~\cite{yao2024minicpm}, Qwen2-VLs~\cite{Qwen2VL}, Qwen2.5-VLs~\cite{qwen2025qwen25technicalreport}, Minimonkey~\cite{huang2024minimonkey}, InternVL2-8B~\cite{chen2024far}, GOT-OCR2.0~\cite{wei2024general}), closed-source VLM models (Qwen-VL-max~\cite{Qwen-VL-max}, GPT-4o~\cite{GPT-4o}), and open-source OCR models (EasyOCR~\cite{EasyOCR}, PP-OCRv4~\cite{PaddleOCR}), Ali-OCR API~\cite{AliOCR}. A YOLOv10~\cite{wang2024yolov10} model is also trained as a strong baseline.

As shown in Table~\ref{tab:full_page_ocr}, with only character-wise slicing and \textit{CalliAlign}, our method surpasses all candidates, including the leading closed-source models GPT-4o and Qwen2.5-VL-Max, achieving over a $16\%$ F1 gain on the easy tier and a $7\%$ gain on the hard, cursive tier. \textit{CalliReader} further extends its advantage over other fine-tuned VLMs after e-IT, particularly on the hard tier, achieving over a $9\%$ lead in F1 score and a $6\%$ reduction in NED, compared to the best competitor, the fine-tuned InternVL2-8B. 


\begin{figure}[t]
    \vspace{-4mm}
    \centering
    \includegraphics[width=\linewidth]{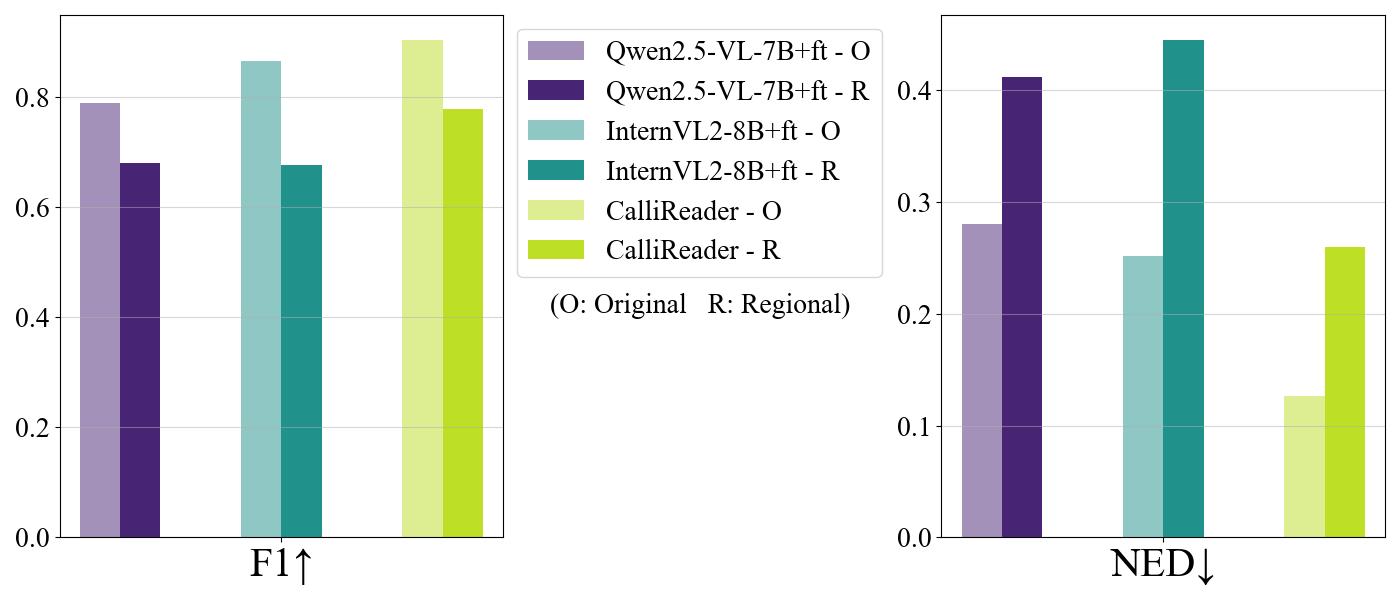}
    \vspace{-9mm}
    \caption{Region hallucination detection. Dark and Light bars denote the model's original and regional recognition capability, indicating the model's resistance and faithfulness to visual content. \textit{CalliReader} exhibits the least drop with leading performance.}
    \label{fig:region_wise_ocr}
    \vspace{-6mm}
\end{figure}

\vspace{-1.5mm}
\subsection{Region Hallucination Identification}
\vspace{-1.5mm}

Calligraphy fragments may cause hallucinations in VLMs due to content inconsistency. This underscores validating the method's robustness. Fig.~\ref{fig:region_wise_ocr} shows all VLMs declining when transitioning from full-page to regional inputs. Notably, \textit{CalliReader} (green bars) maintains the highest performance with the least F1 and NED drops. Unlike conventional image-text fine-tuning, e-IT improves LLM compatibility with pseudo-text embeddings through aligned outputs, faithful to visual contents and reduces hallucinations.


\vspace{-2mm}
\subsection{Contextual VQA}
\vspace{-1.5mm}
We evaluate \textit{CalliReader}'s performance on knowledge selection, bilingual interpretation, and intent analysis. Compared methods include MiniCPM2.6, Qwen2.5-VL-7B, InternVL2-8B, GPT-4o-mini~\cite{OpenAI2024}, along with certain fine-tuned variants (see Table~\ref{tab:entitiy_select}). LoRA and e-IT are tuned on page-level recognition, isolating the impact of CC$^2$ reasoning-related tasks.

Contextual VQA presents a challenging benchmark. Pre- and post-fine-tuning results for InternVL2 and Qwen2.5-VL reveal minimal variation in style, layout, and author recognition but a marked decline in bilingual interpretation and intent analysis (row3-4 vs. row6-7). This degradation stems from single-task LoRA fine-tuning, which weakens instruction-following capabilities.

In contrast, \textit{CalliReader}, with character-wise slicing and \textit{CalliAlign}, surpasses its backbone (row5 v.s. row3). With e-IT, it achieves substantial gains across all metrics (row 5 vs. row 8), reaching 95\% accuracy in authority recognition. Pluggable modules enhance the extraction of visual knowledge, while e-IT via page-level recognition, akin to a reconstruction task, refines pseudo-text token alignment. Without compromising the model's original knowledge and reasoning capacity, this approach strengthens contextual understanding by establishing more precise visual cues.



\begin{table}[t!]
    \centering
    \vspace{-4mm}
    \begin{scriptsize}
    \begin{tabular}{l|c@{\hspace{1em}}c@{\hspace{1em}}c|c@{\hspace{1em}}c}
    \hline
    \multirow{2}{*}{Model} & Style & Layout & Author & Bilingual & Intent \\ \cline{2-6} 
    & \multicolumn{3}{c|}{Acc(\%)↑} & STsim(\%)↑ &TQS(\%)↑\\ \hline
    GPT-4o mini & 34.40 & 38.09& 48.66& 35.52 & 68.93\\
    MiniCPM2.6 & 51.60 & 38.46 & 81.99 & 43.74 & 66.40\\
    InternVL2-8B & 42.00 & 45.66 & 65.91 & 48.37 &70.42\\
    Qwen2.5-VL-7B & 78.53 & 53.85 & 81.23 &  46.29 & 70.80\\
    CalliReader w/o e-IT  & 47.47& 57.44& 79.31 & 52.31 & 74.93\\
    \hline

    InternVL2-8B+ft & 41.60 & 39.70 & 54.61 & 41.47 & 63.95\\
    Qwen2.5-VL-7B+ft & \textbf{78.67} & 51.99 & 83.92 & 28.34 & 14.43\\
    
    \rowcolor{lightgray}CalliReader & 75.33 & \textbf{77.81} & \textbf{95.06} & \textbf{61.14} &\textbf{76.88}\\\hline
\end{tabular}

    \end{scriptsize}
    \vspace{-2.5mm}
    \caption{Contextual VQA (style/layout/author choices, multilingual interpretation, and intent analysis). Without deliberate training, \textit{CalliReader} achieves the best reasoning accuracy.}
    \label{tab:entitiy_select}
    \vspace{-4mm}
\end{table}

\vspace{-1.5mm}
\begin{table}[t]
    \centering
    \setlength{\tabcolsep}{2pt} 
    \small 
    \begin{tabular}{l|ccc|ccc|ccc}
        \hline
        \multirow{2}{*}{Model} & \multicolumn{3}{c|}{SCUT-HCCDoc}  & \multicolumn{3}{c|}{MTHV2} & \multicolumn{3}{c}{OCRBench-cn} \\ \cline{2-10} 
         & P↑ & R↑ & F1↑ & P↑ & R↑ & F1↑ & P↑ & R↑ & F1↑  \\ \hline
        \scriptsize EasyOCR+ft & \scriptsize 0.021 & \scriptsize 0.008  &  \scriptsize 0.011 & \scriptsize 0.053 & \scriptsize 0.015 & \scriptsize 0.023  & \scriptsize 0.004 & \scriptsize 0.004 & \scriptsize 0.003 \\
        \scriptsize PP-OCR+ft & \scriptsize 0.662 & \scriptsize 0.392  &  \scriptsize 0.478 & \scriptsize \textbf{0.819} & \scriptsize 0.609 & \scriptsize 0.696 & \scriptsize 0.280 & \scriptsize 0.203 & \scriptsize 0.222 \\
        \scriptsize InternVL2-8B+ft & \scriptsize 0.528 & \scriptsize 0.513 & \scriptsize 0.467& \scriptsize 0.746 & \scriptsize 0.674 & \scriptsize 0.669& \scriptsize 0.780 & \scriptsize 0.796 & \scriptsize 0.778   \\
        \rowcolor{lightgray} \scriptsize CalliReader & \scriptsize \textbf{0.766} & \scriptsize \textbf{0.578} & \scriptsize \textbf{0.650}  & \scriptsize 0.791 & \scriptsize \textbf{0.706} & \scriptsize \textbf{0.716} & \scriptsize \textbf{0.788} & \scriptsize \textbf{0.796} & \scriptsize \textbf{0.781} \\
        \hline
    \end{tabular}
    \vspace{-2.5mm}
     \captionsetup{font=small}
    \caption{Application on Handwriting (SCUT-HCCDoc), Document (MTHv2), and General (OCRBench-cn) OCR benchmarks. Evaluations prove the superiority of our model.}
    \label{tab:generalization_ocr}
    \vspace{-5mm}
\end{table}

\vspace{-1mm}
\subsection{Recognition Enables Comprehension}
\vspace{-1mm}
Table~\ref{tab:ablation_reason} compares \textit{CalliReader} with state-of-the-art multimodal reasoning models (DeepSeek-Align~\cite{ji2024align}, QvQ-72B-Preview~\cite{qvq-72b-preview}) on contextual VQA multiple-choice tasks. Without character-wise slicing and \textit{CalliAlign}, these models struggle, especially with authorship identification from inscriptions. Insufficient visual content may cause long-chain-of-thought (CoT) VLMs to produce erroneous inferences. Thus, for the CC$^2$ task, precise content recognition is a prerequisite for effective knowledge grounding and contextual understanding.

\vspace{-1.7mm}
\subsection{Generalization Assessment}
\vspace{-1.7mm}
We evaluate \textit{CalliReader}'s generalization using three Chinese benchmarks: SCUT-HCCDoc~\cite{zhang2020scut} (handwriting), MTHV2~\cite{mthv2} (documents), and OCRBench-cn~\cite{ocrbench2023} (real-world scenes). Table~\ref{tab:generalization_ocr} shows that \textit{CalliReader} achieves strong performance across these datasets, maintaining base VLM capabilities. For fairness, all models are untrained on these datasets. Compared to InternVL2-8B+ft, our methods boost performance in general text-centric scenarios. The fine-tuned PP-OCR has better precision on MTHV2 but fails to generalize to other scenes, and EasyOCR can not comprehend these tasks at all, proving conventional OCR tools' fragility on diverse scenes.

\begin{figure}
    \centering
    \vspace{-4mm}
    \begin{minipage}[t]{0.46\linewidth}
        \begin{small}
        \begin{tabular}{lcc}
            \toprule
            Method & L2 & Acc \\
            \midrule
            CalliReader & \textbf{0.096} & \textbf{0.897} \\
            \midrule
            B4→B2 & 0.146 & 0.865 \\
            LN→GN & 0.529 & 0.791 \\\hline
            $+\mathcal{L}_{rat}$ & 0.769 & 0.695 \\
            $+\mathcal{L}_{CRD}$ & 0.317 & 0.813 \\
            \bottomrule
        \end{tabular}
        \end{small}
        \vspace{-3mm}
        \captionsetup{justification=centering}
        \captionof{table}{Ablation study of \textit{CalliAlign} on block size, normalization and losses.}
        \label{tab:ablationcallialign}
    \end{minipage}
    \hspace{5mm}
    \begin{minipage}[t]{0.45\linewidth}
        \begin{small}
        \begin{tabular}{lcc}
            \toprule
            Method & F1 & NED \\
            \midrule
            e-IT+scale & \textbf{0.609} & \textbf{0.516} \\
            \midrule
            e-IT & 0.600 & 0.530 \\
            w/o Align & 0.489 & 0.614 \\\hline
            img & 0.496 & 0.617 \\
            w/o Align & 0.511 & 0.579 \\
            \bottomrule
        \end{tabular}
        \end{small}
        \vspace{-3mm}
        \captionsetup{justification=centering}
        \captionof{table}{Ablation of LLM instruction tuning on embedding or image input.}
        \label{tab:ablationft}
    \end{minipage}
    \vspace{-2mm}
\end{figure}

\begin{figure}
    \centering
    \begin{minipage}[t]{0.46\linewidth}
        \begin{scriptsize}
        \vspace{-10mm}
            \begin{tabular}{@{}c|c@{\hspace{0.7em}}c@{\hspace{0.7em}}c@{\hspace{0.7em}}@{}} 
            \toprule
          \parbox{0.2\linewidth}{\scriptsize Latency (s/page)$\downarrow$} & \scriptsize Easy & \scriptsize Medium & \scriptsize Hard\\
            \midrule
            w/  & 1.48 & \textbf{2.06} & \textbf{33.50} \\
           w/o  & \textbf{1.32} & 2.61 & 41.29 \\
            \bottomrule
        \end{tabular}
        \end{scriptsize}
        \vspace{-3mm}
        \captionsetup{}
        \captionof{table}{Ablation of inference latency. With plug-ins, \textit{CalliReader} surpasses its backbone for less hallucination on the harder tier.}
        \label{tab:ablation_efficiency1}
        \vspace{-6.5mm}
    \end{minipage}
    \hspace{-1mm}
    \begin{minipage}[t]{0.45\linewidth}
        \begin{scriptsize}
        \begin{tabular}{l@{\hspace{0.5em}}|c@{\hspace{0.8em}}c@{\hspace{0.8em}}c@{\hspace{0.8em}}}
            \toprule
            \multirow{2}{*}{Model} & Style & Layout & Author \\ \cline{2-4}  & \multicolumn{3}{c}{Acc(\%)↑} \\ 
            \hline
            DS-Align & 34.17 & 33.87 & 44.27\\
            QvQ-72B & 66.13 & 71.71 & 80.82\\
            \hline
            CalliReader & \textbf{75.33} & \textbf{77.81} & \textbf{95.06} \\
            \bottomrule
        \end{tabular}
        \end{scriptsize}
        \vspace{-3mm}
        \captionsetup{justification=centering}
        \captionof{table}{Performance comparison of multimodal reasoning models on contextual VQA tasks.}
        \label{tab:ablation_reason}
    \end{minipage}
    \vspace{-4mm}
\end{figure}

\begin{figure*}[t]
    \centering
    \vspace{-5.5mm}
    \includegraphics[width=0.999\linewidth]{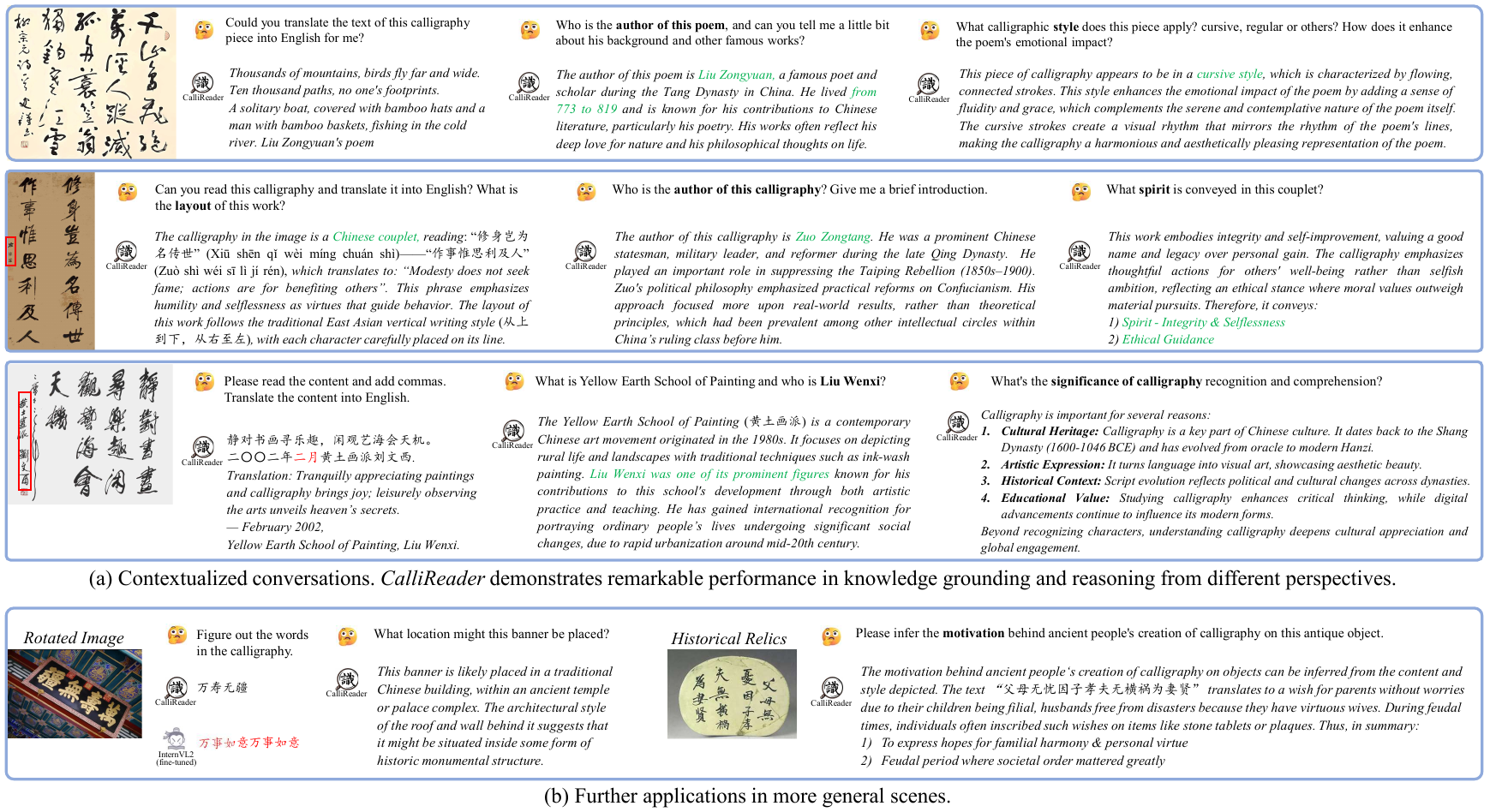}
    \vspace{-7mm}
    \caption{Visualizations of \textit{CalliReader} in free-form, context-aware calligraphic interactions. (a) demonstrates its potential for education and cultural preservation, making calligraphy and its historical context accessible to a global audience. (b) illustrates its visual-text-centric generalization, enabling robust interpretation of rotated and low-quality images. Errors are colored in red. Zoom in for finer details.}
    \label{fig:visualize_final}
    \vspace{-5mm}
\end{figure*}

\vspace{-1mm}
\subsection{Ablation Studies}
\vspace{-1mm}
\noindent\textbf{\textit{CalliAlign} Architecture.}
We ablate \textit{CalliAlign}'s implementations, comparing block sizes (B4: four blocks, B2: 2 blocks), normalization methods (LN: layer-normalization, GN: global-normalization), and loss combinations, including ratio loss $\mathcal{L}_{rat}$ and the contrastive distillation loss $\mathcal{L}_{crd}$~\cite{khosla2020supervised,tian2019contrastive}, both are detailed in the appendix.
We use L2 loss across all settings and evaluate using label-prediction accuracy. Table~\ref{tab:ablationcallialign} shows that a four-block, layer-normalized resampler with L2 loss achieves optimal performance that best fits to pseudo-text embeddings. 


\noindent\textbf{Inference Efficiency.}
We evaluate recognition latency (seconds per page) for \textit{CalliReader} and its backbone to address efficiency concerns. Table~\ref{tab:ablation_efficiency1} shows that with character-wise slicing and \textit{CalliAlign}, \textit{CalliReader} reduces latency in medium and hard tiers by minimizing hallucinations and repetitive outputs. In contrast, the InternVL2 backbone frequently generates repetitive, maximum-length outputs for cursive calligraphy. This also occurs on other compared VLMs, highlighting CC$^2$'s difficulty. Therefore, with a reasonable increase in the parameter count, our plug-in strategy has improved the inference efficiency for recognizing page-level calligraphy artworks.




\noindent\textbf{e-IT and Other Fine-tuning Methods.} We compare embedding instruction tuning (e-IT) and image-text fine-tuning (img) on InternLM2.5-7b-chat, evaluating full-page OCR hard-tier performance using F1 and NED. Table~\ref{tab:ablationft} shows e-IT's better compatibility with pseudo-text embeddings, outperforming e-IT without \textit{CalliAlign} (CA) and image-text fine-tuning with \textit{CalliAlign}. Scaling synthetic data (e-IT+scale) further improves recognition metrics, demonstrating its effectiveness in low-resource scenarios. 

\vspace{-2mm}
\subsection{User Study}
\vspace{-2mm}

We conduct a user study, to demonstrate the difficulty of CC$^2$ task. 142 volunteers, including 18 calligraphy experts, were tasked to recognize 30 randomly selected CalliBench pages. While testees achieved low F1 scores (0.512) and high NED (0.590), \textit{CalliReader} outperformed humans, achieving a 40\% higher F1 score (0.918) and a 50\% reduction in NED (0.092). Details of experiment settings and statistics can be found in the supplementary.

\vspace{-1.5mm}
\subsection{Qualitative Results}
\vspace{-2.5mm}
Fig.~\ref{fig:visualize_final}(a) presents additional examples of \textit{CalliReader} contextualizing rich knowledge, artistry, and emotions from calligraphy pages. It accurately recognizes cursive scripts (first row), varying sizes and layouts (second row for author identification), achieving ``Vision to Knowledge'' through free-form interaction (third row). It also provides biographical insights on calligraphers and authors, interpreting calligraphy within its significance while following user instructions for rich, insightful responses.

Fig.~\ref{fig:visualize_final}(b) further extends \textit{CalliReader} to more general scenes. With character-wise slicing and \textit{CalliAlign}, our method successfully identifies rotated characters, and can decipher historical relics such as stone tabs.



\vspace{-2.5mm}
\section{Discussion and Conclusion}
\vspace{-2.5mm}
In this paper, we proposed \textit{CalliReader}, a novel Vision-Language Model (VLM) specifically designed to interpret knowledge-intensive calligraphic artworks. We also introduced a new benchmark (CalliBench) to address the CC$^2$ task and raise the awareness of scribbled-text-centric, cultural-dimension VQA. Leveraging slicing priors, embedding alignment, and effective fine-tuning, \textit{CalliReader} achieved state-of-the-art performance on CalliBench, surpassing cutting-edge methods and even human professionals. This represents the first comprehensive solution to the task of Chinese Calligraphy Contextualization (CC$^2$). However, there are still large rooms for improvement, especially in processing calligraphy works with cursive writing and complex layouts, as highlighted in Table~\ref{tab:full_page_ocr} and colored in red in Fig.~\ref{fig:visualize_final}. These issues will be addressed in the future.

\newpage
\newpage
{
    \small
    \bibliographystyle{ieeenat_fullname}
    \bibliography{main}
}


\end{document}